\documentclass[runningheads]{llncs}
\usepackage{xcolor}
\usepackage{graphicx}
\usepackage{caption}
\usepackage{subfig}
\usepackage{amssymb}
\usepackage{amsmath}

\usepackage{booktabs}
\usepackage{multirow}
\usepackage{makecell}

\newcommand{\TextUnderscore}{\rule{.4em}{.4pt}}

\usepackage[colorinlistoftodos]{todonotes}

\usepackage[square,sort,comma,numbers]{natbib}

\begin{document}
\title{Representation Disentanglement for Multi-task Learning with application to Fetal Ultrasound}
\titlerunning{Multi-task Representation Disentanglement}
\author{
  Qingjie Meng\inst{1}\and Nick Pawlowski\inst{1}\and Daniel Rueckert\inst{1}\and Bernhard Kainz\inst{1}
}
\authorrunning{Qingjie Meng et al.}
%
\institute{Department of Computing, BioMedIA, Imperial College London, UK\\
\email{q.meng16@imperial.ac.uk}
}
%
\maketitle              
\begin{abstract}
One of the biggest challenges for deep learning algorithms in medical image analysis is the indiscriminate mixing of image properties, \emph{e.g.} artifacts and anatomy. These entangled image properties lead to a semantically redundant feature encoding for the relevant task and thus lead to poor generalization of deep learning algorithms. 
In this paper we propose a novel representation disentanglement method to extract semantically meaningful and generalizable features for different tasks within a multi-task learning framework. Deep neural networks are utilized to ensure that the encoded features are maximally informative with respect to relevant tasks, while an adversarial regularization encourages these features to be disentangled and minimally informative about irrelevant tasks. 
We aim to use the disentangled representations to generalize the applicability of deep neural networks.
We demonstrate the advantages of the proposed method on synthetic data as well as fetal ultrasound images. Our experiments illustrate that our method is capable of learning disentangled internal representations.
It outperforms baseline methods in multiple tasks, especially on images with new properties, \emph{e.g.} previously unseen artifacts in fetal ultrasound. 
\end{abstract}
\section{Introduction}
Image interpretation using convolutional neural networks (CNNs) has been widely and successfully applied to medical image analysis during recent years. 
However, in contrast to human observers,
CNNs exhibit weaknesses of being generalized to tackle previously unseen entangled image properties (\emph{e.g.} shape and texture)~\cite{geirhos2018imagenet}.
In Ultrasound (US), the image property entanglement can be observed when acquisition-related artifacts (\emph{e.g.} shadows) obfuscate the underlying anatomy (see Fig.~\ref{DataPresentation}). A CNN simultaneously learns anatomical features and artifacts features for either anatomy classification or artifacts detection~\cite{meng2018}. As a result, the model trained by images with certain entangled properties (\emph{e.g.} images without acoustic shadows) 
can hardly handle images with new entangled properties which are unseen during training (\emph{e.g.} images with shadows).

\def \fheight {1.1cm}
\begin{figure}[htb]
 \centering
   \begin{tabular}{c@{\hspace{2\tabcolsep}}c@{\hspace{2\tabcolsep}}c@{\hspace{2\tabcolsep}}c@{\hspace{2\tabcolsep}}c@{\hspace{2\tabcolsep}}c@{\hspace{2\tabcolsep}}c@{\hspace{2\tabcolsep}}c}
  \includegraphics[height=\fheight]{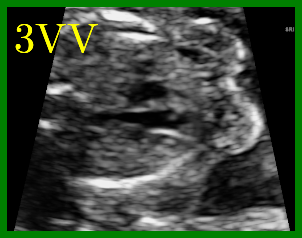} &
  \includegraphics[height=\fheight]{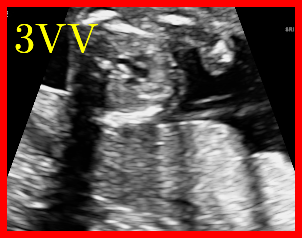}
  &
  \includegraphics[height=\fheight]{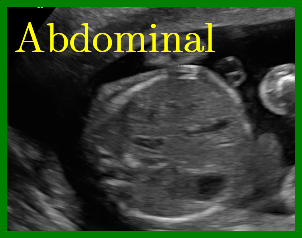} & 
  \includegraphics[height=\fheight]{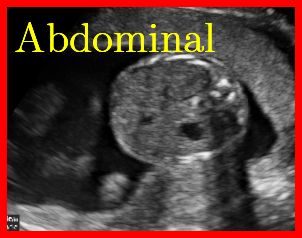}  &
  \includegraphics[height=\fheight]{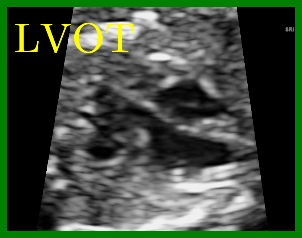} & 
  \includegraphics[height=\fheight]{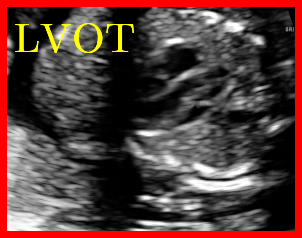}  &
  \includegraphics[height=\fheight]{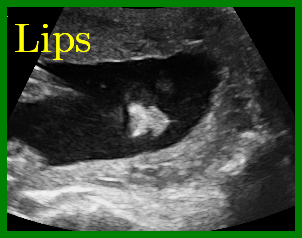}  &
  \includegraphics[height=\fheight]{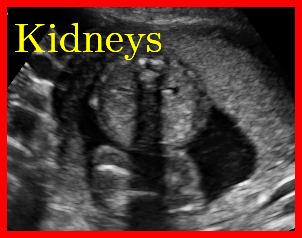}
  \end{tabular}
  \caption{Examples of fetal US data. Green framed images are shadow-free and red framed images contain acoustic shadows.}
  \label{DataPresentation}
\end{figure}
Approaches for representation disentanglement have been proposed in order to learn semantically disjoint internal representations for improving image interpretation~\cite{Kim2018}.
These methods pave a way for improving the generalization of CNNs in a wide range of medical image analysis problems. Specifically for a practical application in this work, we want to disentangle anatomical features from shadow features so that to generalize anatomical standard plane analysis for a better detection of abnormality in early pregnancy.  

\noindent\textbf{Contribution:}  In this paper, we propose a novel, end-to-end trainable representation disentanglement model that can learn distinct and generalizable features through a multi-task architecture with adversarial training. The obtained disjoint features are able to improve the performance of multi-task networks, especially on data with previously unseen properties. We evaluate the proposed model on specific multi-task problems, including shape/background-color classification tasks on synthetic data and standard-plane/shadow-artifacts classification tasks on fetal US data.
Our experiments show that our model is able to disentangle latent representations and, in a practical application, improves the performance for anatomy analysis in US imaging.

\noindent\textbf{Related work:} 
Representation disentanglement has been widely studied in the machine learning literature, ranging from traditional models such as Independent Component Analysis (ICA)~\cite{Hyvarinen2000} and bilinear models~\cite{Tenenbaum2000} to recent deep learning-based models such as InfoGAN~\cite{Chen2016} and $\beta$-VAE~\cite{Higgins2017,Burgess2018}.
Disentangled representations can be utilized to interpret complex interactions of underlying factors within data~\cite{Bengio2013,Chen2017} and enable deep learning models to manipulate relevant information for specific tasks~\cite{Gonzalez-Garcia2018,Liu2018,Hadad2018}.
Particularly related to our work is the work by Mathieu et al.~\cite{Mathieu2016}, which proposed a conditional generative model with adversarial networks to disentangle specific and unspecific factors of variation in deep representations without strong supervision. 
Compared to~\cite{Mathieu2016}, Hadad et al.~\cite{Hadad2018} proposed a simpler two-step method with the same aim. Their network directly utilizes the encoded latent space without assuming the underlying distribution, which can be more efficient for learning various unspecified features. Different from their aim -- disentangling one \emph{specific} representation from \emph{unspecific} factors -- our work focuses on disentangling several \emph{specific} factors.
Further related to our research question is to learn only unspecific invariant features, for example, for domain adaptation~\cite{kamnitsas2017unsupervised}.
However, unlike learning invariant features, which ignores task-irrelevant information~\cite{Bengio2013}, our method aims to preserve information for multiple tasks while enhancing feature generalizability.

In the medical image analysis community, few approaches have focused on disentangling internal factors of representations in discriminative tasks. Ben-Cohen et al.~\cite{Cohen2018} proposed a method to disentangle lesion type from image appearance and use disentangled features to generate more training samples for data augmentation. Their work improves liver lesions classification. In contrast, our work aims to utilize disentangled features for generalization of deep neural networks in medical image analysis.

\section{Method}

\begin{figure}[htb]
 \centering
 \includegraphics[width=\textwidth, trim=4cm 12.6cm 8.4cm 2.2cm, clip]{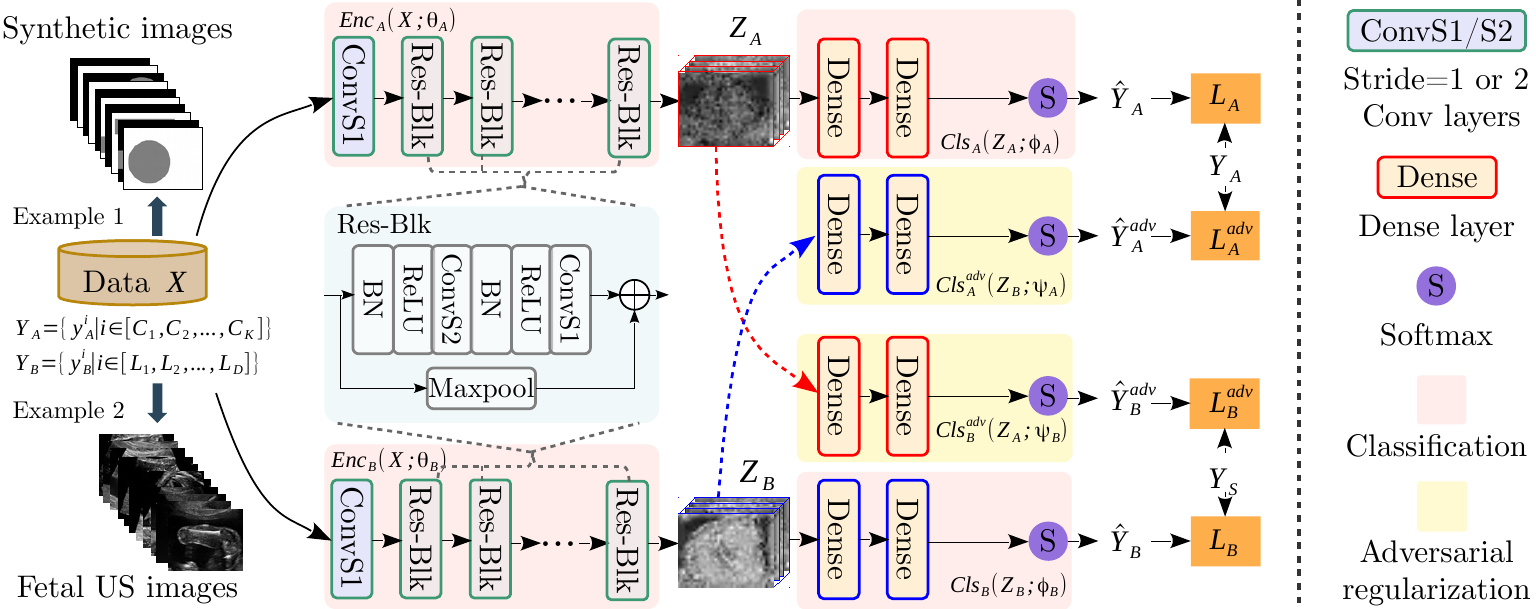}
 \caption{Training framework for the proposed method. Res-Blk refers to residual-blocks. Example 1/2 are two data set examples used in Sect.~\ref{sect:eval}. The classifications enables the encoded features $Z_A, Z_B$ to be maximally informative about related tasks while the adversarial regularization encourages these features to be less informative about irrelevant tasks.}
 \label{flowchart}
\end{figure}

Our goal is to disentangle latent representations $Z$ of the data $X$ into distinct feature sets ($Z_A, Z_B$) that separately contain relevant information for corresponding different tasks ($T_A, T_B$). 
The main motivation of the proposed method is to learn feature sets that are maximally informative about their corresponding task (\emph{e.g.} $Z_A \to T_A$) but minimally representative for irrelevant tasks (\emph{e.g.} $Z_A \to T_B$ ). 
While our approach scales to any number of classification tasks, in this work we focus on two tasks as a proof of concept. 
The proposed method consists of two classification tasks ($T_A, T_B$) with an adversarial regularization. The classification aims to map the encoded features to their relevant class identities, and is trained to maximize $I(Z_A, Y_A)$ and $I(Z_B, Y_B)$. 
The adversarial regularization penalizes the mutual information between the encoded features and their irrelevant class identities, in other words, minimizes $I(Z_A, Y_B)$ and $I(Z_B, Y_A)$. The training architecture of our method is shown in Fig.~\ref{flowchart}.

\noindent\textbf{Classification} is used to learn the encoded features that enable high prediction performance for the class identity of the relevant task. Each of the two classification networks is composed of an encoder and a classifier for a defined task. Given data $X=\{x_i\mid i \in[1,N]\}$, the matching labels are $Y_A=\{y_A^i\mid y_A^i\in\{C_1,C_2,...,C_K\},i\in[1,N]\}$ for $T_A$ and $Y_B=\{y_B^i\mid y_B^i\in\{L_1,L_2,...,L_D\},i\in[1,N]\}$ for $T_B$. $N$ is the number of images and $K,D$ are the number of class identities in each task. Two independent encoders map $X$ to $Z_A$ and $Z_B$ with parameters $\theta_A$ and $\theta_B$ respectively, yielding $Z_A={Enc}_A(X;\theta_A)$ and $Z_B={Enc}_B(X;\theta_B)$. Two classifiers are used to predict class identity for the corresponding task, where $\hat{Y}_A={Cls}_A(Z_A;\phi_A)$ and $\hat{Y}_B={Cls}_B(Z_B;\phi_B)$. $\phi_A$ and $\phi_B$ are the parameters of the corresponding classifiers. We define the the cost functions $\mathcal{L}_A$ and $\mathcal{L}_B$ as the softmax cross-entropy between $Y_A$ and $\hat{Y}_A$ and between $Y_B$ and $\hat{Y}_B$ respectively. The classification loss $\mathcal{L}_{cls}=\mathcal{L}_A+\mathcal{L}_B$ is minimized to train the two encoders and the two classifiers ($\textstyle\min_{\{\theta_A, \theta_B, \phi_A, \phi_B\}} \mathcal{L}_{cls}$) for obtaining $Z_A$ and $Z_B$ that are maximally related to their relevant task.

\noindent\textbf{Adversarial regularization} is used to force the encoded features to be  minimally informative about irrelevant tasks, which results in disentanglement of internal representations. The adversarial regularization is implemented by using an adversarial network for each task as shown in Fig.~\ref{flowchart}. These adversarial networks are utilized to map the encoded features to class identity of the irrelevant task, yielding $\hat{Y}_A^{adv}={Cls}_A^{adv}(Z_B;\psi_A)$ and $\hat{Y}_B^{adv}={Cls}_B^{adv}(Z_A;\psi_B)$. Here, $\psi_A$ and $\psi_B$ are the parameters of the corresponding adversarial networks. By referring to $\mathcal{L}_A^{adv}$ and $\mathcal{L}_B^{adv}$ as the softmax cross-entropy between $Y_A$ and $\hat{Y}_A^{adv}$ and between $Y_B$ and $\hat{Y}_B^{adv}$, the adversarial loss is defined as $\mathcal{L}_{adv}=\mathcal{L}_A^{adv}+\mathcal{L}_B^{adv}$. During training, the adversarial networks are trained to minimize $\mathcal{L}_{adv}$ while two encoders and two classifiers are trained to maximize $\mathcal{L}_{adv}$ ($\textstyle\min_{\{\psi_A, \psi_B\}} \max_{\{\theta_A, \theta_B, \phi_A, \phi_B\}} \mathcal{L}_{adv}$). This competition between the encoders/classifiers and the adversarial networks encourages the encoded features to be invalid for irrelevant tasks. 

By combining the two classifications with the adversarial regularization, the whole model is optimized iteratively during training. The training objective for optimizing the two encoders and the two classifiers can be written as 
\begin{equation}\label{Loss_dc}
\textstyle\min_{\{\theta_A, \theta_B, \phi_A, \phi_B\}} {\{\mathcal{L}_A+\mathcal{L}_B-\lambda*(\mathcal{L}_A^{adv}+\mathcal{L}_B^{adv})\}}, \; \lambda > 0.
\end{equation}
Here, $\lambda$ is the trade-off parameter of the adversarial regularization. The training objective for the optimization of the adversarial networks thus follows as 
\begin{equation}\label{Loss_adv}
\textstyle\min_{\{\psi_A, \psi_B\}} \{\mathcal{L}_A^{adv}+\mathcal{L}_B^{adv}\}.
\end{equation}

\noindent\textbf{Network architectures:} ${Enc}_A(X;\theta_A)$ and ${Enc}_B(X;\theta_B)$ both consist of six residual-blocks implemented as proposed in~\cite{pawlowski2017state} to reduce the training error and to support easier network optimization.
${Cls}_A(Z_A;\phi_A)$ and ${Cls}_B(Z_B;\phi_B)$ both contain two dense layers with $256$ hidden units. 
The adversarial networks ${Cls}_A^{adv}(Z_B;\psi_A)$ and ${Cls}_B^{adv}(Z_A;\psi_B)$ have the same architecture as ${Cls}_A(Z_A;\phi_A)$ and ${Cls}_B(Z_B;\phi_B)$ respectively.

\noindent\textbf{Training:} 
Our model is optimized for $400$ epochs and $\lambda$ is chosen heuristically and independently for each data set using validation data.
For more stable optimization~\cite{Hadad2018}, in each iteration, we train the encoders and classifiers once, followed by five training steps of the adversarial networks. 
Similar to~\cite{Hadad2018}, we use the Adam optimizer ($\text{beta}=0.9$, $\text{learning rate}=10^{-5}$) to train the encoders and classifiers based on Eq.~\ref{Loss_dc}, and use Stochastic Gradient Descent (SGD) with momentum optimizer ($\text{momentum}=0.9$, $\text{learning rate}=10^{-5}$) to update the parameters of the adversarial networks in Eq.~\ref{Loss_adv}. We apply L2 regularization ($\text{scale}=10^{-5}$) to all weights during training to prevent over-fitting. The batch size is 50 and the images in each batch have been randomly flipped as data augmentation. 
Our model is trained on a Nvidia Titan X GPU with 12 GB of memory.

\section{Evaluation and Results}
\label{sect:eval}

\begin{figure}[!ht]
 \centering
  \subfloat{\includegraphics[height=2.2cm, trim=6.7cm 5cm 1.6cm 1.4cm, clip]{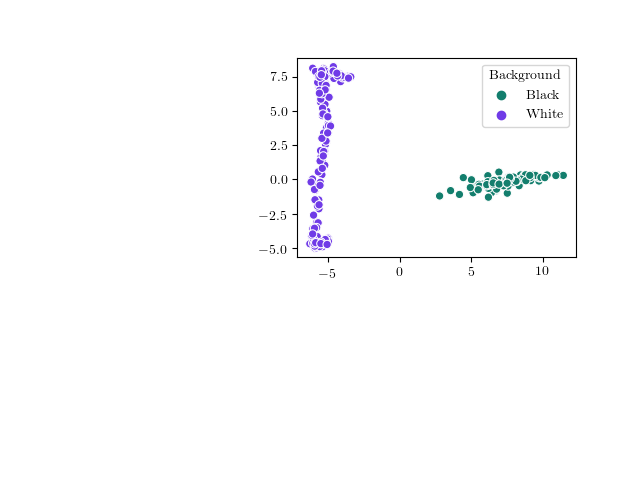}} \hfill 
  \subfloat{\includegraphics[height=2.2cm, trim=6.7cm 5cm 1.6cm 1.4cm, clip]{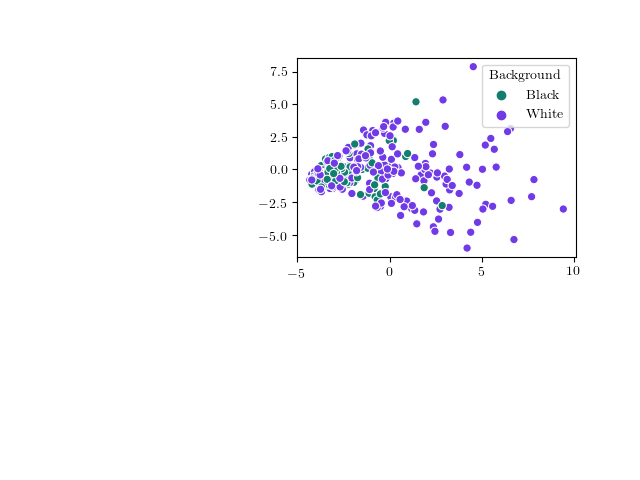}} \hfill 
  \subfloat{\includegraphics[height=2.2cm, trim=6.7cm 5cm 1.6cm 1.4cm, clip]{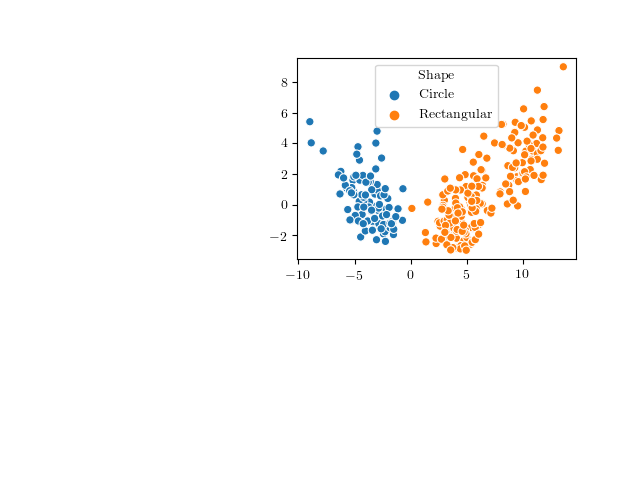}} \hfill 
  \subfloat{\includegraphics[height=2.2cm, trim=6.7cm 5cm 1.6cm 1.4cm, clip]{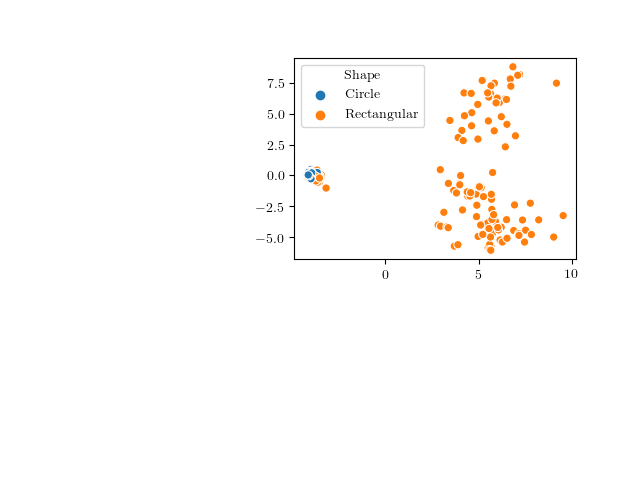}} \hfill 
  \\
 \setcounter{subfigure}{0}
   \subfloat[$Z_A \to T_A$]{\includegraphics[height=2.2cm, trim=6.7cm 5cm 1.6cm 1.4cm, clip]{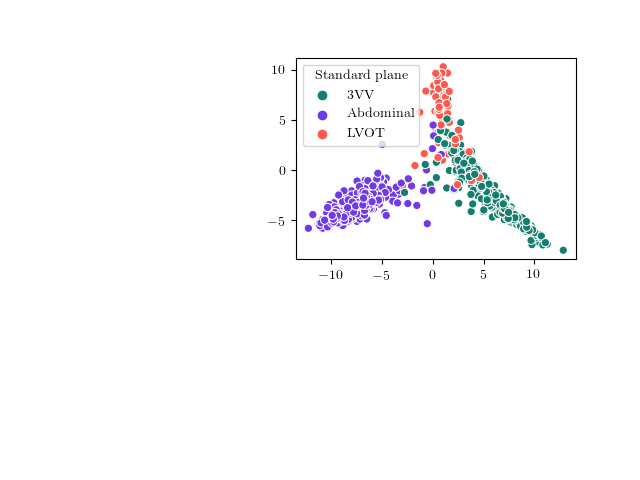}} \hfill 
  \subfloat[$Z_B \to T_A$]{\includegraphics[height=2.2cm, trim=6.7cm 5cm 1.6cm 1.4cm, clip]{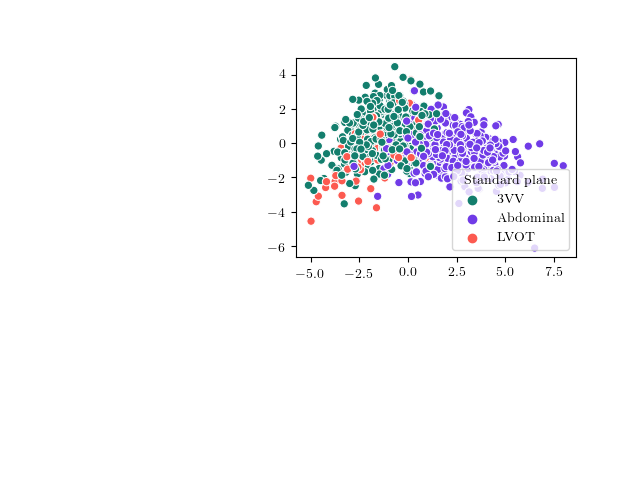}} \hfill
  \subfloat[$Z_B \to T_B$]{\includegraphics[height=2.2cm, trim=6.7cm 5cm 1.6cm 1.4cm, clip]{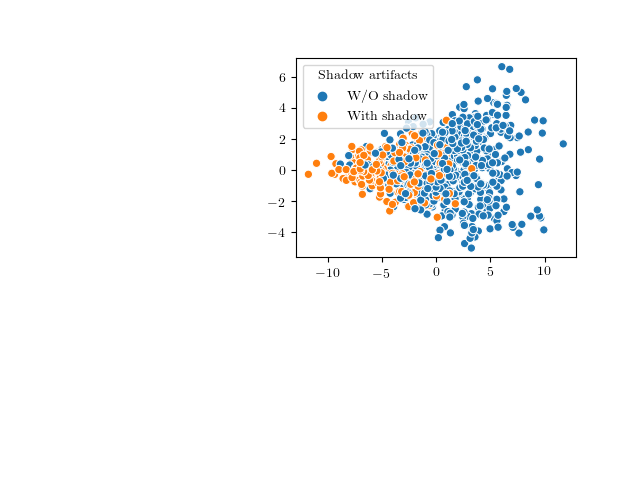}} \hfill 
  \subfloat[$Z_A \to T_B$]{\includegraphics[height=2.2cm, trim=6.7cm 5cm 1.6cm 1.4cm, clip]{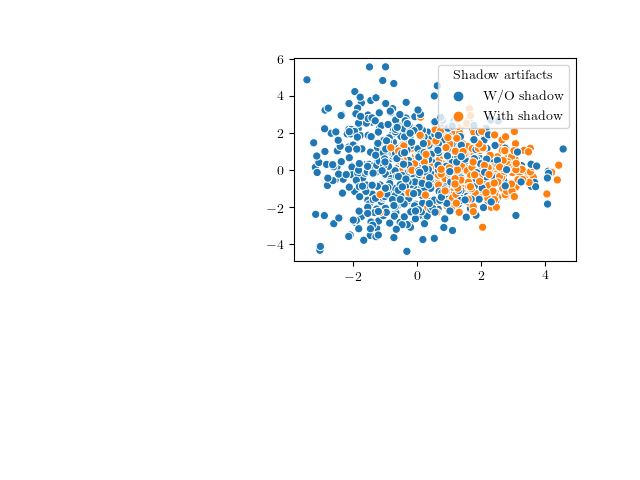}} \hfill 
  \\
  \caption{Visualization of the embedded data on the penultimate dense layer. The top row shows embedded synthetic test data while the bottom row shows embedded fetal US Test\TextUnderscore{}seen data. (a, c) are the results of using encoded features for relevant tasks, \emph{e.g.} $Z_A$ for $T_A$ and $Z_B$ for $T_B$; separated clusters are desirable here. (b, d) are the results of using encoded features for irrelevant tasks, namely, $Z_A$ for $T_B$ and $Z_B$ for $T_A$; mixed clusters are desirable in this case.}
  \label{visualization}
\end{figure}

\noindent\textbf{Evaluation on synthetic data:}
We use synthetic data as a proof of concept example to verify our model. This data set contains a randomly located gray circle or rectangle on a black or white background. We split the data into $1200/300/300$ images for train/validation/test and these images consist of circles on white background, rectangles on black background and rectangles on white background. To keep the balance between image properties in the training split, we use circle:rectangle=1:1 and black:white=7:5. 
In this case, $T_A$ is a background color classification task and $T_B$ is the a shape classification task. We implement our model as outlined in Sec.2
and choose $\lambda=0.01$.
We evaluate our model on the test data. The experimentation illustrates that the encoded features successfully identify the class identities of the relevant task (\emph{e.g.} $Z_A \to T_A:$ $\text{OA}_{acc}=100\%$, $Z_B \to T_B:$ $\text{OA}_{acc}=99.67\%$) but fail to handle irrelevant task (\emph{e.g.} $Z_A \to T_B:$ $\text{OA}_{acc}=62\%$, $Z_B \to T_A:$ $\text{OA}_{acc}=59.67\%$). 
Here, $\text{OA}_{acc}$ is the overall accuracy.
To show the utility of the proposed method on images with previously unseen entangled properties, we additionally compare the shape classification performance of our model and a baseline (our model without the adversarial regularization) on images with a previously unseen entangled properties (circles on black background). The proposed model achieves $\text{OA}_{acc}=99\%$ and  outperforms the baseline which achieves $\text{OA}_{acc}=10\%$.
We use PCA to examine the learned embedding space at the penultimate dense layer of the classifiers. The top row of Fig.~\ref{visualization} illustrates that the extracted features is able to identify class identities for relevant tasks (see (a,c)) but unable to predict correct class identities for irrelevant tasks (see (b,d).


\noindent\textbf{Evaluation on fetal US data:}
We verify the applicability of our method on fetal US data. Here, we refer to an anatomical standard plane classification task as $T_A$ and an acoustic shadow artifacts classification task as $T_B$. We want to learn the corresponding disentangled features $Z_A$ for all anatomical information, separated from $Z_B$ containing only information about shadow artifacts. $Y_A$ is the label for different anatomical standard planes while $Y_B^i=0$ and $Y_B^i=1$ are the labels of the shadow-free class and the shadow-containing class respectively.

\noindent\textbf{Data set:} The fetal US data set contains $8.4k$ images sampled from 4120 2D US fetal anomaly screening examinations with gestational ages between 18$-$22 weeks.  
These sequences consist of eight standard planes defined in the UK FASP handbook~\cite{screening2015}, including three vessel view (3VV), left ventricular outflow tract (LVOT), abdominal (Abd.), four chamber view (4CH), femur, kidneys, lips and right ventricular outflow tract (RVOT), and are classified by expert observers as shadow-containing (W\TextUnderscore{}S) or shadow-free (W/O\TextUnderscore{}S) (Fig.~\ref{DataPresentation}). We split the data as shown in Table.~\ref{datasplit}.
Train, Validation and Test\TextUnderscore{}seen are separate data sets. Test\TextUnderscore{}seen contains the same entangled properties (but different images) as used for the training data set, while LVOT(W\TextUnderscore{}S) and Artifacts(OTHS) contain new combinations of entangled properties.
\begin{table}[t]
\centering
\caption{Data split. ``Others" contains standard planes 4CH, femur, kidneys, lips and RVOT. Test\TextUnderscore{}seen, LVOT(W\TextUnderscore{}S) and Artifacts(OTHS) are used for testing.}
\label{datasplit}
\begin{tabular}{*7c}
\toprule
~~~~~~ &
~~~~~~ &
Train  &
Validation  &
Test\TextUnderscore{}seen  &
LVOT(W\TextUnderscore{}S)  &
Artifacts(OTHS)  \\
\cmidrule{2-7}
3VV &
W/O\TextUnderscore{}S (W\TextUnderscore{}S)  &
180 (320)  &
50 (50)  &
334 (41)  &
- (-)  &
- (-)  \\
LVOT  &
W/O\TextUnderscore{}S (W\TextUnderscore{}S)  &
500 (-)  &
50 (-)  &
79 (-)  &
- (418)  &
- (-)  \\
Abd.  &
W/O\TextUnderscore{}S (W\TextUnderscore{}S)  &
125 (375)  &
50 (50) &
190 (220)  &
- (-) &
- (-)  \\
Others &
W/O\TextUnderscore{}S (W\TextUnderscore{}S)  &
- (-)  &
- (-)  &
- (-)  &
- (-)  &
3159 (2211)  \\
\bottomrule
\end{tabular}
\end{table}


\noindent\textbf{Evaluation approach:} We refer to \textit{Std plane only} as the networks for standard plane classification only (consists of ${Enc}_A$ and ${Cls}_A$), and \textit{Artifacts only} as the networks for shadow artifacts classification only (consists of ${Enc}_B$ and ${Cls}_B$). \textit{$\text{Proposed}_{w/o\TextUnderscore{}{adv}}$} refers to the proposed method without the adversarial regularization and \textit{Proposed} is our method in Fig.~\ref{flowchart}.

The proposed method is implemented as outlined in Sec.2 choosing $\lambda=0.1$. ${Cls}_A(Z_A;\phi_A)$ contains three dense layers with $256/256/3$ hidden units while ${Cls}_B(Z_B;\phi_B)$ contains two dense layers with $256/2$ hidden units. We choose a bigger network capacity for ${Cls}_A(Z_A;\phi_A)$ by assuming that anatomies have more complex structures than shadows to be learned.

Table.~\ref{cls} shows that our method improves the performance of standard plane classification by $16.08\%$ and $13.19\%$ on {Test\TextUnderscore{}seen} when compared with the \textit{Std plane only} and the $\text{Proposed}_{w/o\TextUnderscore{}{adv}}$ method (see $\text{OA}_{acc}$ in Col.5). 
It achieves minimal improvement (\textit{Artifacts only}: $+0.35\%$ and $\text{Proposed}_{w/o\TextUnderscore{}{adv}}$: $+0.31\%$ classification accuracy) for shadow artifacts classification (see $\text{OA}_{acc}$ in Col.8).
We also demonstrate the utility of the proposed method on images with previously unseen entangled properties. Table.~\ref{cls} shows that the proposed method achieves $73.68\%$ accuracy of standard plane classification on LVOT(W\TextUnderscore{}S) ($\sim 36\%$ higher than other comparison methods) while it performs similar to other methods on Artifacts(OTHS) for shadow artifacts classification.

\begin{table}[t]
\centering
\caption{The classification accuracy ($\%$) of different methods for the standard classification ($T_A$) and shadow artifacts classification ($T_B$) on Test\TextUnderscore{}seen data set and data sets with unseen entangled properties (LVOT(W\TextUnderscore{}S) and Artifacts(OTHS)). ``Proposed" uses encoded features for relevant tasks, namely, $Z_A\to T_A$ and $Z_B\to T_B$. ``$\text{Proposed}_{irr\TextUnderscore{}task}$" uses encoded features for irrelevant tasks, namely, $Z_A\to T_B$ and $Z_B\to T_A$. $\text{OA}_{acc}$ is the overall accuracy.}
\label{cls}
\begin{tabular}{*{10}c}
\toprule
Col.1  &
Col.2  &
Col.3  &
Col.4  &
Col.5  &
Col.6  &
Col.7  &
Col.8  &
Col.9  &
Col.10  \\
\midrule
\multirow{2}{*}{Methods}  &
\multicolumn{7}{c}{Test\TextUnderscore{}seen}  &
\multirow{2}{*}{\thinspace \makecell{LVOT \\ (W\TextUnderscore{}S)} \thinspace}  &
\multirow{2}{*}{\makecell{Artifacts \\ (OTHS)}} \\
\cmidrule(lr){2-8}
~~~~~  &
3VV \thinspace  &
LVOT \thinspace  &
Abd. \thinspace  &
$\text{OA}_{acc}$  &
W/O\TextUnderscore{}S \thinspace  &
W\TextUnderscore{}S \thinspace  &
$\text{OA}_{acc}$   &
~~~~~~  &
~~~~~~  \\
\cmidrule(lr){2-5}\cmidrule(lr){6-8}
Std plane only  &
60.80  &
96.59  &
67.09  &
78.36  &
-  &
-  &
-  &
34.93  &
-  \\
Artifacts only  &
-  &
-  &
-  &
-  &
77.94  &
80.46  &
78.70  &
-  &
69.26  \\
$\text{Proposed}_{w/o\TextUnderscore{}{adv}}$ &
63.73  &
97.80  &
78.48  &
81.25  &
78.77  &
77.78  &
78.74  &
37.56  &
69.50  \\
\textbf{Proposed} & 
93.87  &
97.56  &
81.01  &
\textbf{94.44}  &
87.89  &
58.62  &
\textbf{79.05}  &
\textbf{73.68}  &
68.49  \\
$\text{Proposed}_{irr\TextUnderscore{}task}$   &
39.20  &
83.90  &
82.28  &
64.35  &
68.49  &
81.99  &
72.57  &
-  &
-  \\
\bottomrule
\end{tabular}
\end{table}

We evaluate the performance of disentanglement by using the encoded features for the irrelevant task on Test\TextUnderscore{}seen, \emph{e.g.} $Z_A \to T_B$ and $Z_B \to T_A$. Here, $Z_A$ and $Z_B$ are encoded features of the proposed method. $\text{Proposed}_{irr\TextUnderscore{}task}$ in Table.~\ref{cls} indicates that $Z_B$ contains much less anatomical information for standard plane classification ($\text{OA}_{acc}=94.44\%$ in proposed vs. $\text{OA}_{acc}=64.35\%$ in $\text{Proposed}_{irr\TextUnderscore{}task}$), while $Z_A$ contains less shadow features information ($\text{OA}_{acc}=79.05\%$ in proposed vs. $\text{OA}_{acc}=72.57\%$ in $\text{Proposed}_{irr\TextUnderscore{}task}$).
We additionally use PCA to show the embedded test data on the penultimate dense layer.
The bottom row in Fig.~\ref{visualization} shows that encoded features are more capable of classifying class identities in the relevant task than the irrelevant task (\emph{e.g.} (a) vs. (d)).

\noindent\textbf{Discussion: } 
Acoustic shadows are caused by anatomies which block the propagation of sound waves or by destructive interference. With this dependency between anatomy and artifacts, separating shadow features from anatomical features may lead to decreased performance of artifacts classification (Table.\ref{cls}, Col.7, Proposed). However, this separation enables feature generalization so that the model is less limited to certain image formation and able to tackle new combinations of entangled properties (Table.\ref{cls}, Col.9, Proposed).  
Generalization of supervised neural networks can also be achieved by extensive data collection across domains and in a limited way by artificial data augmentation. Here, we propose an alternative through feature disentanglement, which requires less data collection and training effort. 
Fig.~\ref{visualization} shows PCA plots for the penultimate dense layer. Observing entanglement in earlier layers reveals that disentanglement occurs in this very last layer. This is due to the definition of our loss functions and is partly influenced by the dense layers interpreting the latent representation for classification. 
Finally, perfect representation disentanglement is likely infeasible because image features are rarely totally isolated in reality. 
In this paper we have shown that even imperfect disentanglement is able to provide great benefits for artifact-prone image classification in medical image analysis. 

\section{Conclusion}
In this paper, we propose a novel disentanglement method to extract generalizable features within a multi-task framework.
In the proposed method, classification tasks lead to encoded features that are maximally informative with respect to these tasks while the adversarial regularization forces these features to be minimally informative about irrelevant tasks, which disentangles internal representations. Experimental results on synthetic and fetal US data show that our method outperforms baseline methods for multiple tasks, especially on images with entangled properties that are unseen during training.
Future work will explore the extension of this framework to multiple tasks beyond classification.

\subsubsection*{Acknowledgments.} We thank the Wellcome Trust IEH Award [102431], Nvidia (GPU donations) and Intel.

\bibliographystyle{unsrtnat}

\begin{thebibliography}{18}
\providecommand{\natexlab}[1]{#1}
\providecommand{\url}[1]{\texttt{#1}}
\expandafter\ifx\csname urlstyle\endcsname\relax
  \providecommand{\doi}[1]{doi: #1}\else
  \providecommand{\doi}{doi: \begingroup \urlstyle{rm}\Url}\fi

\bibitem[Geirhos et~al.(2018)Geirhos, Rubisch, Michaelis, Bethge, Wichmann, and
  Brendel]{geirhos2018imagenet}
Robert Geirhos, Patricia Rubisch, Claudio Michaelis, Matthias Bethge, Felix~A.
  Wichmann, and Wieland Brendel.
\newblock Imagenet-trained cnns are biased towards texture; increasing shape
  bias improves accuracy and robustness.
\newblock \emph{arXiv:1811.12231}, 2018.

\bibitem[Meng et~al.(2019)Meng, Sinclair, Zimmer, Hou, Rajchl, Toussaint,
  Oktay, Schlemper, Gomez, Housden, Matthew, Rueckert, Schnabel, and
  Kainz]{meng2018}
Qingjie Meng, Matthew Sinclair, Veronika Zimmer, Benjamin Hou, Martin Rajchl,
  Nicolas Toussaint, Ozan Oktay, Jo~Schlemper, Alberto Gomez, James Housden,
  Jacqueline Matthew, Daniel Rueckert, Julia~A Schnabel, and Bernhard Kainz.
\newblock Weakly supervised estimation of shadow confidence maps in fetal
  ultrasound imaging.
\newblock \emph{IEEE transactions on medical imaging}, 2019.
\newblock ISSN 0278-0062.

\bibitem[Kim and Mnih(2018)]{Kim2018}
Hyunjik Kim and Andriy Mnih.
\newblock Disentangling by factorising.
\newblock \emph{CoRR}, arXiv/1802.05983, 2018.

\bibitem[Hyv\"{a}rinen and Oja(2000)]{Hyvarinen2000}
A.~Hyv\"{a}rinen and E.~Oja.
\newblock Independent component analysis: Algorithms and applications.
\newblock \emph{Neural Netw.}, 13\penalty0 (4-5):\penalty0 411--430, May 2000.
\newblock ISSN 0893-6080.

\bibitem[Tenenbaum and Freeman(2000)]{Tenenbaum2000}
Joshua~B. Tenenbaum and William~T. Freeman.
\newblock Separating style and content with bilinear models.
\newblock \emph{Neural Comput.}, 12\penalty0 (6):\penalty0 1247--1283, June
  2000.
\newblock ISSN 0899-7667.

\bibitem[Chen et~al.(2016)Chen, Duan, Houthooft, Schulman, Sutskever, and
  Abbeel]{Chen2016}
Xi~Chen, Yan Duan, Rein Houthooft, John Schulman, Ilya Sutskever, and Pieter
  Abbeel.
\newblock Infogan: Interpretable representation learning by information
  maximizing generative adversarial nets.
\newblock In \emph{NeurIPS'16}, pages 2180--2188, USA, 2016. Curran Associates
  Inc.
\newblock ISBN 978-1-5108-3881-9.

\bibitem[Higgins et~al.(2017)Higgins, Matthey, Pal, Burgess, Glorot, Botvinick,
  Mohamed, and Lerchner]{Higgins2017}
Irina Higgins, Loic Matthey, Arka Pal, Christopher Burgess, Xavier Glorot,
  Matthew Botvinick, Shakir Mohamed, and Alexander Lerchner.
\newblock beta-vae: Learning basic visual concepts with a constrained
  variational framework.
\newblock In \emph{ICLR'17}, 2017.

\bibitem[Burgess et~al.(2018)Burgess, Higgins, Pal, Matthey, Watters,
  Desjardins, and Lerchner]{Burgess2018}
Christopher~P. Burgess, Irina Higgins, Arka Pal, Lo{\"{\i}}c Matthey, Nick
  Watters, Guillaume Desjardins, and Alexander Lerchner.
\newblock Understanding disentangling in {\(\beta\)}-vae.
\newblock \emph{arXiv:1804.03599}, 2018.

\bibitem[Bengio et~al.(2013)Bengio, Courville, and Vincent]{Bengio2013}
Yoshua Bengio, Aaron Courville, and Pascal Vincent.
\newblock Representation learning: A review and new perspectives.
\newblock \emph{IEEE Trans. Pattern Anal. Mach. Intell.}, 35\penalty0
  (8):\penalty0 1798--1828, August 2013.
\newblock ISSN 0162-8828.

\bibitem[Chen et~al.(2017)Chen, Kingma, Salimans, Duan, Dhariwal, Schulman,
  Sutskever, and Abbeel]{Chen2017}
Xi~Chen, Diederik~P Kingma, Tim Salimans, Yan Duan, Prafulla Dhariwal, John
  Schulman, Ilya Sutskever, and Pieter Abbeel.
\newblock Variational lossy autoencoder.
\newblock In \emph{ICLR'17}, 2017.

\bibitem[Gonzalez-Garcia et~al.(2018)Gonzalez-Garcia, van~de Weijer, and
  Bengio]{Gonzalez-Garcia2018}
Abel Gonzalez-Garcia, Joost van~de Weijer, and Yoshua Bengio.
\newblock Image-to-image translation for cross-domain disentanglement.
\newblock In \emph{NeurIPS'18}, pages 1287--1298. Curran Associates, Inc.,
  2018.

\bibitem[Liu et~al.(2018)Liu, Liu, Yeh, and Wang]{Liu2018}
Alexander~H. Liu, Yen-Cheng Liu, Yu-Ying Yeh, and Yu-Chiang~Frank Wang.
\newblock A unified feature disentangler for multi-domain image translation and
  manipulation.
\newblock In \emph{NeurIPS}, pages 2590--2599. Curran Associates, Inc., 2018.

\bibitem[Hadad et~al.(2018)Hadad, Wolf, and Shahar]{Hadad2018}
Naama Hadad, Lior Wolf, and Moni Shahar.
\newblock A two-step disentanglement method.
\newblock In \emph{CVPR'18}, 2018.

\bibitem[Mathieu et~al.(2016)Mathieu, Zhao, Zhao, Ramesh, Sprechmann, and
  LeCun]{Mathieu2016}
Michael~F Mathieu, Junbo~Jake Zhao, Junbo Zhao, Aditya Ramesh, Pablo
  Sprechmann, and Yann LeCun.
\newblock Disentangling factors of variation in deep representations using
  adversarial training.
\newblock In \emph{NeurIPS'16}, pages 5040--5048, 2016.

\bibitem[Kamnitsas et~al.(2017)Kamnitsas, Baumgartner, Ledig, Newcombe,
  Simpson, Kane, Menon, Nori, Criminisi, Rueckert, and
  Glocker]{kamnitsas2017unsupervised}
Konstantinos Kamnitsas, Christian Baumgartner, Christian Ledig, Virginia
  Newcombe, Joanna Simpson, Andrew Kane, David Menon, Aditya Nori, Antonio
  Criminisi, Daniel Rueckert, and Ben Glocker.
\newblock Unsupervised domain adaptation in brain lesion segmentation with
  adversarial networks.
\newblock In \emph{International conference on information processing in
  medical imaging}, pages 597--609. Springer, 2017.

\bibitem[Ben{-}Cohen et~al.(2018)Ben{-}Cohen, Mechrez, Yedidia, and
  Greenspan]{Cohen2018}
Avi Ben{-}Cohen, Roey Mechrez, Noa Yedidia, and Hayit Greenspan.
\newblock Improving {CNN} training using disentanglement for liver lesion
  classification in {CT}.
\newblock \emph{arXiv:1811.00501}, 2018.

\bibitem[Pawlowski et~al.(2017)Pawlowski, Ktena, Lee, Kainz, Rueckert, Glocker,
  and Rajchl]{pawlowski2017state}
Nick Pawlowski, S.~Ira Ktena, Matthew Lee, Bernhard Kainz, Daniel Rueckert, Ben
  Glocker, and Martin Rajchl.
\newblock Dltk: State of the art reference implementations for deep learning on
  medical images.
\newblock \emph{arXiv:1711.06853}, 2017.

\bibitem[{NHS}(2015)]{screening2015}
{NHS}.
\newblock \emph{Fetal anomaly screening programme: programme handbook June
  2015}.
\newblock Public Health England, 2015.

\end{thebibliography}

\end{document}